\pdfoutput=1

\documentclass[11pt]{article}

\usepackage{eacl2023}

\usepackage{times}
\usepackage{latexsym}
\usepackage{hyperref}
\usepackage[frozencache=true,cachedir=minted-cache]{minted}
\usepackage{array}
\usepackage{xcolor}
\usepackage{graphicx}
\usepackage{multirow}
\usepackage{booktabs}

\usepackage[T1]{fontenc}

\usepackage[utf8]{inputenc}

\usepackage{microtype}

\def\kogito{\texttt{\textbf{k}ogito}}
\newcommand\thinker{\raisebox{-2pt}{\includegraphics[width=0.9em]{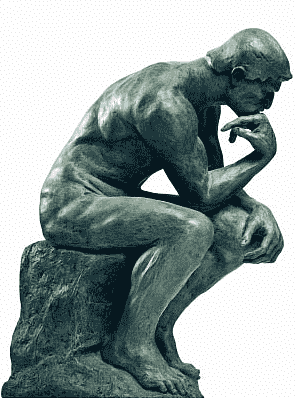}}}

\newcommand\eg{\textit{e.g.}}
\newcommand\ie{\textit{i.e.}}

\title{\thinker{} \kogito{}: A Commonsense Knowledge Inference Toolkit}

\author{Mete Ismayilzada \\
  EPFL \\
  \texttt{mahammad.ismayilzada@epfl.ch} \\\And
  Antoine Bosselut \\
  EPFL \\
  \texttt{antoine.bosselut@epfl.ch} \\}

\begin{document}
\maketitle
\begin{abstract}
In this paper, we present \kogito{}, an open-source tool for generating commonsense inferences about situations described in text. \kogito{} provides an intuitive and extensible interface to interact with natural language generation models that can be used for hypothesizing commonsense knowledge inference from a textual input. In particular, \kogito{} offers several features for targeted, multi-granularity knowledge generation. These include a standardized API for training and evaluating knowledge models, and generating and filtering inferences from them. We also include helper functions for converting natural language texts into a format ingestible by knowledge models --- intermediate pipeline stages such as knowledge head extraction from text, heuristic and model-based knowledge head-relation matching, and an ability to define and use custom knowledge relations. We make the code for \kogito{} available at \href{https://github.com/epfl-nlp/kogito}{https://github.com/epfl-nlp/kogito}  along with thorough documentation at \href{https://kogito.readthedocs.io}{https://kogito.readthedocs.io}.
\end{abstract}

\section{Introduction} \label{introduction}
In recent years, large-scale language models \cite{Radford2018ImprovingLU,devlin-etal-2019-bert,NEURIPS2020_1457c0d6} trained on massive amounts of text have been conceptualized as implicit knowledge bases that encode knowledge about the world \cite{petroni-etal-2019-language,roberts-etal-2020-much}. As they are trained to receive natural language inputs, these models can be prompted to generate text that expresses a fact. Leveraging this property, \textit{knowlege models} train on knowledge graph tuples (triplets of \textit{head entity}, \textit{relation}, \textit{tail entity}) and learn to express knowledge encoded in the parameters of language models when provided with a \textit{head entity} and \textit{relation} \cite{bosselut-etal-2019-comet, Hwang2021COMETATOMIC2O,Da2020AnalyzingCE,West2022SymbolicKD}. 

The success of these \textit{knowledge models} has inspired the field to deploy them in various downstream use-cases such as generating figurative language \cite{chakrabarty-etal-2020-generating}, producing sarcastic responses \cite{chakrabarty-etal-2020-r}, designing plots for stories \cite{Ammanabrolu2021AutomatedSV} and text-based games \cite{dambekodi-etal-2020}, and developing persona-grounded dialogue agents \cite{majumder-etal-2020-like}. 
Given the prevalence of applications that benefit from augmenting NLP systems with commonsense inferences, we present a novel commonsense \textbf{K}n\textbf{O}wled\textbf{G}e \textbf{I}nference \textbf{TO}olkit, \kogito{}, that standardizes commonsense inference generation from knowledge models. 
To the best of our knowledge, \kogito{} is the first library that facilitates access to knowledge models through an easy-to-use, customizable interface. In particular, we make the following contributions:


\begin{enumerate}
    \item A Python package\footnote{\url{https://pypi.org/project/kogito/}} for knowledge inference with a customizable and extensible API.
    \item A module to perform commonsense inference with a library of pretrained models, including GPT-2 \citep{radford2019language}, GPT-3 \citep{NEURIPS2020_1457c0d6} and COMET \citep{Hwang2021COMETATOMIC2O}.
    \item A standardized interface to train, evaluate and predict with knowledge models.
    \item Modules to extract relevant candidates for commonsense inference (\ie, head extraction) with support for customization and extension.
    \item Modules to match relevant relations to extracted head entities (\ie, relation matching) with support for customization and extension.
    \item A module to filter commonsense inferences based on their contextual relevance using commonsense fact linkers \cite{silin2022comfact}
    \item Functionality to define novel knowledge relations on top of the built-in ATOMIC2020 \cite{Hwang2021COMETATOMIC2O} and ConceptNet \cite{Speer2013ConceptNet5A} relation sets.
    \item Extensive documentation with User Guides and API Reference.\footnote{\url{https://kogito.readthedocs.io/}}
\end{enumerate}

\noindent The library is released under the Apache 2.0 License. We provide a demo video\footnote{\url{https://www.youtube.com/watch?v=rFGzDrLCx00}} for our library along with a live demo app.\footnote{\url{https://kogito.live}} Below, we outline the commonsense inference challenges addressed by this tool, its core design, and walk through its major components in more detail.

\section{Challenges of Commonsense Inference}
\label{sec:challenges}

While many works use \textit{knowledge models} as commonsense inference engines to augment natural language inputs, no work has formalized the pipeline for producing such inferences. Here, we outline the challenges of effectively setting up this pipeline.

\paragraph{Head Extraction} 
Head extraction (\ie, finding relevant concepts to produce commonsense inferences about) is a consistent challenge when using knowledge models. Typically, these inferences must be produced for more fine-grained textual units than full contexts \citep{Bosselut2019DynamicKG}. For instance, to understand figurative language, \citet{chakrabarty-etal-2020-generating} extract concepts from similes such as \textit{"Love is like a unicorn"}. Commonsense inferences are generated about entities such as \textit{"unicorn"} (\eg, unicorns are \textit{rare}, \textit{beautiful}, etc.), allowing them to produce literal interpretations of this figurative language: \textit{"Love is rare"}. This use case motivates a need for fine-grained text extraction functionality in our tool. In Section \ref{sec:head-extraction}, we outline our approach to address this challenge.

\paragraph{Relation Matching} 
To generate commonsense inferences, knowledge models typically take as input a \textit{(head, relation)} pair and produce a \textit{tail} (\ie, the commonsense inference about the \textit{head} entity). Following this convention, once we have extracted candidate \textit{heads} from a given text, they must be paired with relevant relations to produce valid commonsense inferences. For example, a head entity such as ``go to mall'' should not be paired with an \texttt{ObjectUse} relation as it is unlikely to produce a valid (and practical) commonsense inference. Consequently, a brute-force approach of matching all relations to presented head entities would be inadequate for most use cases. 
Current works often circumvent this challenge by manually selecting only a subset of available knowledge relations. As part of \kogito{}, we implement various heuristic and model-based matching schemes to address this challenge, while also providing users with the ability to define their own matching mechanisms. We discuss these implementations in Section \ref{sec:relation-matching}.

\paragraph{Inference Generation \& Filtering} 
Once a list of relevant \textit{(head, relation)} pairs is produced, we run these examples through a knowledge model to generate tail entities about these examples. However, many of these generated inferences may not be relevant to the original context, particularly for extracted head entities that have been de-contextualized. \kogito{} leverages a model-based approach \cite{silin2022comfact} to filter out irrelevant commonsense generations.  While other works re-implement pipelines for performing these steps, 
\kogito{} offers an all-in-one solution. 


\section{\kogito{}: A Pipeline for Commonsense Inference} 
\label{sec:core-design}

\kogito{} is a pipeline for commonsense inference from text and supports various steps to specialize and customize inference behaviour. At full functionality, given a text input, \kogito{} extracts relevant \textit{knowledge heads} from textual inputs, and matches these heads to plausible \textit{knowledge relations}, thereby constructing an incomplete \textit{knowledge graph} of (\textit{head}, \textit{relation}) prompts. Then, this partial graph is input to a \textit{knowledge model} to generate tails to complete the graph. Finally, these commonsense inferences (comprised of the \textit{head}, \textit{relation}, and \textit{tail}) are filtered based on their relevance to the initial context. Below we provide a simple example of how this module can be used to generate commonsense inferences for the example \textit{"PersonX becomes a great basketball player"}:
\begin{minted}[fontsize=\scriptsize]{python}
from kogito.models.bart.comet import COMETBART
from kogito.inference import CommonsenseInference

# Load pre-trained model from HuggingFace
model = COMETBART
        .from_pretrained("mismayil/comet-bart-ai2")

# Initialize inference module
csi = CommonsenseInference()

# Run inference
text = "PersonX becomes a great basketball player"
kgraph = csi.infer(text, model)

# Save output knowledge graph to JSON file
kgraph.to_jsonl("kgraph.json")
\end{minted}
\noindent The resulting knowledge graph from the code above contains inferences such as \textit{"PersonX needs to practice a lot"} and \textit{"PersonX is athletic"}. Various parts of this pipeline can be customized and modified, allowing users to define their own modules. In the following sections, we discuss \kogito's core design, as well as the \textit{head extraction}, \textit{relation matching}, and \textit{inference filtering} components of the pipeline. More details on these configuration options can be found in the kogito documentation.
\section{Data Representation}

To allow for standardization and ease of maintenance, \kogito{} defines an interface to represent core concepts such as a \textit{knowledge tuple}, a \textit{commonsense knowledge graph}, and a \textit{knowledge model}.

\paragraph{Commonsense Knowledge Tuple} Commonsense knowledge graphs \cite{Speer2013ConceptNet5A,Hwang2021COMETATOMIC2O} and knowledge models \cite{bosselut-etal-2019-comet} typically represent instances of knowledge as tuples of 3 elements: \textit{(head, relation, tail)}. The \textit{head} entity refers to the subject of a piece of knowledge (\eg, objects such as \textit{hammer}; events such as \textit{"PersonX becomes a great basketball player"}). A \textit{relation} provides an implicit question about the \textit{head} (\eg, \texttt{CapableOf} $\rightarrow$ \textit{what is this head entity capable of?}; \texttt{xNeed} $\rightarrow$ \textit{What does PersonX need before this event occurs?}). Finally, \textit{tail} entities provide an answer option (among potentially many) to the relation with respect to the head (\eg, \textit{put nail in wood}; \textit{to practice hard}). We often refer to the \textit{tail} as the commonsense inference about the \textit{head}.

Following this convention, we define a class with these elements and an additional two classes for knowledge \textit{head} and \textit{relation} representation. While knowledge \textit{heads} and \textit{tails} can be arbitrary text, we use predefined \textit{relations} from the ATOMIC2020 \cite{Hwang2021COMETATOMIC2O} and ConceptNet \cite{Speer2013ConceptNet5A} knowledge graphs.\footnote{\kogito{} also supports defining new custom relations and using them to generate commonsense inferences (\S\ref{sec:custom-relations})} 
Below is an example of defining a knowledge tuple in \kogito{}:

\begin{minted}[fontsize=\scriptsize]{python}
from kogito.core.head import KnowledgeHead
from kogito.core.knowledge import Knowledge
from kogito.core.relation import X_NEED

head = KnowledgeHead("PersonX becomes a great
                      "basketball player")
kg = Knowledge(head=head, relation=X_NEED,
               tails=["practice hard"])
\end{minted}

\paragraph{Knowledge Graph} In addition to individual knowledge tuples, we also define a knowledge graph as a collection of knowledge tuples. In \kogito{}, a knowledge graph serves as the standardized input object to (and output from) knowledge models,
and has a simple API to manipulate knowledge instances collectively. In particular, a knowledge graph can be used to easily iterate over, read, and write a collection of knowledge instances to and from various files, and perform set-like operations on multiple knowledge graphs. These operations require a notion of equality, so we define two knowledge instances to be equal if they have the same head, relation and tail. Below is an example of defining and manipulating knowledge graphs:
\begin{minted}[fontsize=\scriptsize]{python}
from kogito.core.knowledge import KnowledgeGraph

# Read from csv
kgraph1 = KnowledgeGraph
            .from_csv("sample_graph1.csv",
                      sep="|", header=None)

# Read from jsonl (list of json objects)
kgraph2 = KnowledgeGraph
            .from_jsonl("sample_graph2.jsonl",
                        head_attr="source",
                        relation_attr="rel",
                        tails_attr="targets")

# Union 
# kgraph1.union(kgraph2)
kgraph3 = kgraph1 + kgraph2

# Intersection
# kgraph1.intersection(kgraph2)
kgraph3 = kgraph1 & kgraph2 

# Difference
# kgraph1.difference(kgraph2)
kgraph3 = kgraph1 - kgraph2 

# Write to jsonl
kgraph3.to_jsonl("sample_graph3.jsonl")
\end{minted}

\paragraph{Knowledge Model} 
Knowledge models conceptually accept as input a \textit{(head, relation)} pair and output an inferred knowledge tail.
However, these models can sometimes expect a subtly different formats for these inputs and outputs. To increase the extensibility and interoperability of our tool, so that users can easily substitute one knowledge model for another, we define a model-agnostic abstraction over possible types of knowledge models.
Consequently, knowledge models inherit from an abstract interface that defines core abstract methods, 
which users can implement to port new knowledge models into \kogito{}. The \textit{knowledge model} interface provides methods to train, generate from, evaluate these models, as well as save and load them. 
The input dataset for training, generating, or evaluating is given as a \textit{knowledge graph} and the output dataset (in the case of generation) is returned as a \textit{knowledge graph}. 
%

\kogito{} currently offers the following knowledge models: COMET-BART and COMET-GPT2 from \citet{Hwang2021COMETATOMIC2O}, GPT2-zeroshot \cite{radford2019language}, and GPT3-zeroshot \cite{NEURIPS2020_1457c0d6}.
%
\noindent 
Pre-trained COMET models can be loaded either from HuggingFace\footnote{\url{https://huggingface.co/models}} by name or from local disk by model path. The GPT-3 model requires an API key. Each model method supports customization of various model-specific hyperparameters. \kogito{} currently evaluates models using the following metrics: \textit{BLEU} \cite{papineni-etal-2002-bleu}, \textit{ROUGE} \cite{lin-2004-rouge}, \textit{METEOR} \cite{lavie-agarwal-2007-meteor}, \textit{CIDEr} \cite{7299087} and \textit{BERTScore} \cite{https://doi.org/10.48550/arxiv.1904.09675}.

\paragraph{Pipeline Design} In the following sections, we discuss the \textit{head extraction}, \textit{relation matching}, and \textit{inference filtering} components of this pipeline. We note that we provide a ``dry-run'' mode which allows for faster iteration on head extraction and relation matching by skipping the inference generation portion of \kogito's pipeline. More details on these configuration options can be found in the kogito docs.
\section{Head Extraction} 
\label{sec:head-extraction}

Head extraction refers to finding relevant chunks of a text in a sequence that can serve as knowledge heads (\ie, the concepts commonsense inferences should be generated about). For example, given a text input \textit{"PersonX becomes a great basketball player"}, we might be interested in generating inferences for the full sentence, but also about entities such as \textit{"basketball player"}, \textit{"basketball"}, or potentially \textit{"become player"}. For different applications, different sets of head entities might be appropriate for generating inferences. Consequently, \kogito{} allows the user to customize this behaviour and define arbitrary head extraction methods.\footnote{\url{https://tinyurl.com/head-extraction}}

At the same time, by default, \kogito{} comes with a few standard head extraction methods. These built-in methods segment sentences, and then extract noun phrases (NP) and verb phrases (VP) using dependency parses produced from spaCy.\footnote{\url{https://spacy.io/}} Extracted heads are deduplicated using string matching and passed onto the next stage of the pipeline, \textit{relation matching}. We note that the \textit{head extraction} stage itself is optional and the user can also provide a dedicated list of heads to \kogito, which would replace the pre-processing of head entities.
\section{Relation Matching} 
\label{sec:relation-matching}
Not all relations that a knowledge model is trained with will be relevant to each extracted head. For example, a head entity, \textit{"hammer"}, would ideally be match to a relation such as \texttt{AtLocation}, while a relation such as \texttt{xWants} (\ie, what does this head entity want) would not be matched. Similarly, \textit{"PersonX becomes a basketball player"} might be matched to a relation such as \texttt{xIntent} (\ie, what is the intent of the main persona in the head entity), while a relation such as \texttt{UsedFor} (\ie, what is the head entity used for) would yield an incoherent inference. In this next stage, \kogito{} matches relations to the given head input so that the resulting \textit{(head, relation)} pair constitutes a sensible and plausible prompt for the knowledge model.

\kogito{} supports relation matching as a pre-processing step before generating inferences. Suitable relation matches may be subjective depending on the use case, so \kogito{} supports specifying subsets of relations and creation of custom relation matching modules developed by the user.\footnote{\url{https://tinyurl.com/relation-matching}}

In addition, \kogito{} also provides native relation matching algorithms. These relation matchers follow the categorization of relations set out by \citet{Hwang2021COMETATOMIC2O}, where relations were mapped into three categories: \textit{Physical}, \textit{Social} and \textit{Event} types. Following this standard, we design relation matchers that identify a given head with whether it should be connected to the \textit{Physical}, \textit{Social} or \textit{Event} categories, and match all relations in these categories to the head entity. Below, we describe three relation matching methods provided as part of \kogito{}'s core library:

\paragraph{Base Matcher} Every relation defined for a knowledge graph is matched to the head entities. This matcher is particularly useful if the user pre-defines a set of acceptable known relations or if they define new relations for their use case (\S\ref{sec:custom-relations}).

\paragraph{Heuristic matcher} The heuristic relation matcher matches extracted head entities that are noun phrases to ATOMIC2020 \textit{Physical} relations and extracted head entities that are sentences or verb phrases to \textit{Social} and \textit{Event} relations. In our example, \textit{"PersonX becomes a great basketball player"}, an extracted verb phrase such as \textit{"become player"} would be matched to the \textit{Social} and \textit{Event} relations, while the extracted noun phrase \textit{"basketball player"} would be matched to the \textit{Physical} relations. 

    
\subsection{Model-based relation matching} 
\label{ssec:model_based}
The above matchers do not consider the semantic meaning of the head entities when matching them to relations. We also define model-based matchers that learn which heads and relations would be good matches. Relation matching is modeled as a classification problem. A head entity is given as input, and the model must determine the relation groups that match: \textit{Physical, Social} and \textit{Event}.


\begin{table}[t]
    \centering
    \begin{tabular}{rccc} 
     \toprule
     \textbf{Dataset} & \textbf{$n_{train}$} & \textbf{$n_{test}$} & \textbf{Overlap}  \\
     \toprule
     Original & 36,940 & 6,559 & 0.80 / 0.81 \\
     \midrule
     $D_4$ & 40,395 & 1,192 & 0.30 / 0.36 \\
     $D_2$ & 40,516 & 1,071 & 0.20 / 0.27 \\
     $D_0$ & 40,777 & 810 & 0.00 / 0.11 \\
     \bottomrule
    \end{tabular}
    \caption{Summary of Relation Matching datasets. The overlap column reports the degree of overlap with / without stopwords included.}
    \label{tab:rm-dataset}
\end{table}

\paragraph{Dataset} We use the ATOMIC2020 knowledge graph to train and evaluate the model-based relation matchers. First, we construct a dataset where the inputs are head entities and the label space corresponds to the three relation groups. If a head entity in the knowledge graph is connected to a relation from a particular group, we treat that relation group as a positive label for the head entity. As relations from multiple relation groups may be connected to a head entity, this labeling yields a multi-label prediction problem.

To evaluate the performance of our relation matchers (and test their generalization so they may be applicable to a broad scope of use cases), we split our dataset into both an in-distribution (ID) and an out-of-distribution (OOD) evaluation sample set. For the ID test set, we use the original ATOMIC2020 development set. For the OOD test set, we combine the train and test set of ATOMIC2020 and resplit this joint dataset while minimizing the word overlap between the train and test set. 
More specifically, we prepare 3 sets of (train, test) splits called $D_0$, $D_2$ and $D_4$ where $n$ in $D_n$ is defined as the maximum number of times a word in a particular test set example can occur in the training dataset (excluding stopwords). A bigger $n$ indicates more overlap between these two sets. In $D_0$, the test set does not have any overlapping non-stopwords with the training set. Finally, we ensure that the resulting test set is balanced over each relation group. Table \ref{tab:rm-dataset} provides the summary of the constructed datasets and Table \ref{tab:sample-n1} lists some examples from the $D_0$ dataset. 
\begin{table}[t]
    \resizebox{\linewidth}{!}{
    \centering

    \begin{tabular}{ ccc }
        \toprule
        \textbf{Split} & {\textbf{Head Entity}} & {\textbf{Labels}} \\
        \toprule
        Train & PersonX acts funny & \textit{event, social} \\
        Train & accordion & \textit{physical}\\
        Train & big investment & \textit{event} \\
        Test & agenda & \textit{physical} \\
        Test & PersonX wreaks havoc & \textit{event, social} \\
        Test & PersonX motivates PersonY & \textit{social} \\
        \bottomrule
    \end{tabular}
    }
    \caption{Samples from resplit train and test set $D_0$}
    
    \label{tab:sample-n1}
\end{table}
\paragraph{Models} We report results for fine-tuned models using different pretrained embeddings: GloVe \cite{pennington-etal-2014-glove}, BERT \cite{devlin-etal-2019-bert} and DistilBERT \cite{https://doi.org/10.48550/arxiv.1910.01108}. The GloVe model uses the technique of \citet{shen-etal-2018-baseline} with average pooling over 100 dimensional GloVe embeddings and a projection layer on top. The BERT and DistilBERT models are finetuned on the task with a projection layer to predict the label.\footnote{All models are trained using binary cross-entropy loss and the Adam optimizer \citep{adam2015} for 20 (for SWEM models) and 3 (for BERT and DistilBERT models) epochs with a batch size of 64.} These models are provided with \kogito{}, and can be selected to match relations to head inputs.

In Table \ref{tab:rm-models}, we report the train, ID test and OOD test F1 scores for these models using different training datasets $D_n$, allowing users to understand their relative benefits and trade-offs. 

\begin{table}[t]
    \resizebox{\linewidth}{!}{
    \centering
    \begin{tabular}{ llrrr } 
        \toprule
        \textbf{Data} &\textbf{Model} & \textbf{Train F1} & \textbf{ID F1} & \textbf{OOD F1} \\
        
        \toprule
        \multirow{5}{*}{$D_4$} & Base & 0.68 & 0.82 & 0.62 \\
        & Heuristic & 0.84 & 0.80 & 0.69 \\
        & GloVe  & 0.90 & 0.91 & 0.82 \\
        & DistilBERT  & \textbf{0.97} & 0.91 & 0.85 \\
        & BERT  & \textbf{0.97} & \textbf{0.91} & \textbf{0.86} \\
        \midrule
        \multirow{5}{*}{$D_2$} & Base & 0.68 & 0.82 & 0.61 \\
        & Heuristic & 0.84 & 0.80 & 0.69 \\
        & GloVe  & 0.89 & 0.90 & 0.81 \\
        & DistilBERT  & \textbf{0.97} & 0.93 & 0.85 \\
        & BERT  & \textbf{0.97} & \textbf{0.94} & \textbf{0.86} \\
        \midrule
        \multirow{5}{*}{$D_0$} & Base & 0.68 & 0.82 & 0.63 \\
        & Heuristic & 0.84 & 0.80 & 0.73 \\
        & GloVe  & 0.89 & 0.90 & 0.76 \\
        & DistilBERT  & \textbf{0.97} & \textbf{0.93} & 0.84 \\
        & BERT  & \textbf{0.97} & 0.91 & \textbf{0.85} \\
        \bottomrule
    \end{tabular}
    }
    \caption{Relation matcher performance on datasets $D_n$}
    \label{tab:rm-models}
\end{table}

\section{Inference Filtering} \label{ssec:inference-filtering}
By default, the commonsense inference module returns all generated tails without any filtering applied. However, many of these resulting inferences may be irrelevant to the initial context, particularly for extracted heads that have been de-contextualized. Given most users may only be interested in relevant subsets of these commonsense inferences, \kogito{} provides a separate module to determine the relevance of the given \textit{knowledge tuples} with respect to the initial \textit{context} from which it was extracted. In our running example, \textit{"PersonX becomes a great basketball player"}, an extracted head entity \textit{"player"} may yield contextually-irrelevant inferences such as \textit{"player plays video games"} and \textit{"player is at a soccer match"}, which would be filtered.

To filter inferences, \kogito{} comes with the off-the-shelf DeBERTa-based commonsense fact linking model from \citet{silin2022comfact}, which achieved a state-of-the-art average 72.5\% F1 across multiple benchmarks.  However, our setting is different from the one evaluated in \citet{silin2022comfact} as we evaluate generated commonsense inferences (rather than ones from an existing KB) for contextual relevance. To evaluate how well our method transfers to this new setting, we perform an expert study on the performance of the inference filtering model with respect to the knowledge generated from a knowledge model such as COMET. We randomly select 50 instances from the test split of ROC-ATOMIC dataset \citet{silin2022comfact} where each instance is composed of a \textit{context} and a \textit{fact} as a knowledge tuple \textit{(head, relation, tail)}. We then run the default \kogito{} inference pipeline (with full head extraction and heuristic relation matching) on the \textit{heads} which produces several inferences per head instance. We select 100 results randomly from the output of the previous step and apply our inference filtering model. Finally, we ask a human expert to annotate each instance with the true relevance label of the fact and find that our model achieves a 75\% F1 on the knowledge model generated inferences. We also offer a modular interface to define and plug in new filtering models in the future.

\section{Defining New Relations} 
\label{sec:custom-relations}
In previous knowledge modeling papers \citep{bosselut-etal-2019-comet,Hwang2021COMETATOMIC2O}, the set of relations that can be used in prompts is limited by the knowledge graph used to to train the knowledge model (\eg, ATOMIC2020). However, a user may want to generate inferences for new dimensions of knowledge, define their own custom relations for them, and produce commonsense inferences based on these new properties. However, if there are no large KGs that use this schema, training a suitable knowledge model would pose a challenge. 

\kogito{} provides this functionality by implementing the approach of \citet{West2022SymbolicKD}, which allows a user to prompt large language models for knowledge using custom relations and has been shown to generate high-quality knowledge. Specifically, a user defines an instance of a \textit{knowledge relation} class, a verbalizer function that describes how to convert the new relation into a natural language prompt (with a head and tail), and an instruction prompt to GPT-3. At inference time, the user provides a list of sample knowledge tuples that use the new relation. These tuples are verbalized using the verbalizer function and provided to the GPT-3 model along with the instruction prompt. Below, we illustrate this process with an example where a new relation, \texttt{xWishes}, which describes person's wishes, is defined using the sample code:
\begin{minted}[fontsize=\scriptsize]{python}
from kogito.core.relation import (KnowledgeRelation,
                                  register_relation)

def x_wishes_verbalizer(head, **kwargs):
   # index will be passed from the model
   # so that we can enumerate samples
   # which helps with inference
   index = kwargs.get("index")
   index_txt = f"{index}" if index is not None \
                                        else ""
   return f"Situation {index_txt}: {head}."
           "As a result, PersonX wishes"

X_WISHES = KnowledgeRelation("xWishes",
            verbalizer=x_wishes_verbalizer,
            prompt="How does this situation affect"
                    " each character's wishes?")
register_relation(X_WISHES)
\end{minted}
Then, to use this new relation for inference, the user can provide a sample knowledge graph (\ie, a prompt filled with example tuples using this relation), and a head such as \textit{"PersonX makes a huge mistake"} to generate inferences about. Below, we show how such a sample knowledge graph could be verbalized into a prompt for GPT-3:\\
\fbox{\begin{minipage}{20em}
\noindent \textcolor{magenta}{How does the situation affect the character's wishes?}\\
\textcolor{blue}{Situation 1: John is at a party. As a result, John wishes to drink beer and dance}\\
\textcolor{blue}{Situation 2: Terry bleeds a lot. As a result, Terry wishes to see a doctor}\\
\textcolor{blue}{Situation 3: Eileen works as a cashier. As a result, Eileen wishes to be a store manager}\\
\textcolor{blue}{Situation 4: James gets dirty. As a result, James wishes to clean up}\\
\textcolor{blue}{Situation 5: Janice stays up all night studying. As a result, Janice wishes to sleep all day}\\
\textcolor{orange}{Situation 6: Isaac makes a huge mistake. As a result, Isaac wishes...}
\end{minipage}} 

\noindent The result of prompting GPT-3 with the above text is returned as the generated tail inference for the given head. Using this approach, users can instantiate a prompt defining a new relation, and use large language models to produce inferences for it. 
\section{Conclusion \& Future Work} 
\label{sec:conclusion}
In this system description, we presented \kogito{}, a toolkit for generating commonsense inferences for open-world text using knowledge models. \kogito{} provides a foundational, customizable, and extensible interface for inference generation from knowledge models, and supports preprocessing and manipulation utilities such as head extraction, relation matching, and relation definition. 

Future work may include improved head extraction, such as semantic head extraction (\eg, paraphrased noun phrase extraction, etc.), new relation matching methods that more rigorously trade off performance and latency, support for new knowledge models trained on other knowledge graphs (\eg, ANION; \citealp{jiang-etal-2021-im}), and multimodal inputs such as images.


\section*{Acknowledgements}
We thank Silin Gao, Deniz Bayazit, Beatriz Borges, Antoine Masanet, and other members of the EPFL NLP lab for their feedback on earlier iterations of this library. Significant portions of the model training and evaluation code for this tool have been adapted from the codebase\footnote{\url{https://github.com/allenai/comet-atomic-2020}} of \citet{Hwang2021COMETATOMIC2O}. Antoine Bosselut gratefully acknowledges the support of Innosuisse under PFFS-21-29, the EPFL Science Seed Fund, the EPFL Center for Imaging, Sony Group Corporation, and the Allen Institute for AI.

\section*{Ethical Considerations}

\kogito{} is a library that uses knowledge models such as COMET \citep{bosselut-etal-2019-comet} to generate commonsense inferences from text. These knowledge models are seeded with pretrained language models and subsequently finetuned on knowledge graphs so that they may generate knowledge in the structure of the finetuning KG. Consequently, \kogito{} could reflect harmful behaviors exhibited by language models and knowledge graphs
that are used to train the knowledge models in its library. For example, language models have been shown to encode biases about race, gender, and many other demographic attributes \citep{sheng-etal-2020-towards,ethical-social-risks}. They can also generate toxic outputs when prompted in overt \citep{wallace-etal-2019-universal}, but also seemingly innocuous \citep{Gehman2020RealToxicityPromptsEN}, ways. We encourage users of this library to consider the same precautions they would apply to other language models and methods that use noisy knowledge sources. 

\bibliographystyle{acl_natbib}
\bibliography{literature,anthology}

\begin{thebibliography}{32}
\expandafter\ifx\csname natexlab\endcsname\relax\def\natexlab#1{#1}\fi

\bibitem[{Ammanabrolu et~al.(2021)Ammanabrolu, Cheung, Broniec, and
  Riedl}]{Ammanabrolu2021AutomatedSV}
Prithviraj Ammanabrolu, Wesley Cheung, William Broniec, and Mark~O. Riedl.
  2021.
\newblock Automated storytelling via causal, commonsense plot ordering.
\newblock In \emph{AAAI}.

\bibitem[{Bosselut et~al.(2021)Bosselut, Bras, and
  Choi}]{Bosselut2019DynamicKG}
Antoine Bosselut, Ronan~Le Bras, and Yejin Choi. 2021.
\newblock Dynamic neuro-symbolic knowledge graph construction for zero-shot
  commonsense question answering.
\newblock In \emph{Proceedings of the 35th AAAI Conference on Artificial
  Intelligence (AAAI)}.

\bibitem[{Bosselut et~al.(2019)Bosselut, Rashkin, Sap, Malaviya, Celikyilmaz,
  and Choi}]{bosselut-etal-2019-comet}
Antoine Bosselut, Hannah Rashkin, Maarten Sap, Chaitanya Malaviya, Asli
  Celikyilmaz, and Yejin Choi. 2019.
\newblock \href {https://doi.org/10.18653/v1/P19-1470} {{COMET}: Commonsense
  transformers for automatic knowledge graph construction}.
\newblock In \emph{Proceedings of the 57th Annual Meeting of the Association
  for Computational Linguistics}, pages 4762--4779, Florence, Italy.
  Association for Computational Linguistics.

\bibitem[{Brown et~al.(2020)Brown, Mann, Ryder, Subbiah, Kaplan, Dhariwal,
  Neelakantan, Shyam, Sastry, Askell, Agarwal, Herbert-Voss, Krueger, Henighan,
  Child, Ramesh, Ziegler, Wu, Winter, Hesse, Chen, Sigler, Litwin, Gray, Chess,
  Clark, Berner, McCandlish, Radford, Sutskever, and
  Amodei}]{NEURIPS2020_1457c0d6}
Tom Brown, Benjamin Mann, Nick Ryder, Melanie Subbiah, Jared~D Kaplan, Prafulla
  Dhariwal, Arvind Neelakantan, Pranav Shyam, Girish Sastry, Amanda Askell,
  Sandhini Agarwal, Ariel Herbert-Voss, Gretchen Krueger, Tom Henighan, Rewon
  Child, Aditya Ramesh, Daniel Ziegler, Jeffrey Wu, Clemens Winter, Chris
  Hesse, Mark Chen, Eric Sigler, Mateusz Litwin, Scott Gray, Benjamin Chess,
  Jack Clark, Christopher Berner, Sam McCandlish, Alec Radford, Ilya Sutskever,
  and Dario Amodei. 2020.
\newblock \href
  {https://proceedings.neurips.cc/paper/2020/file/1457c0d6bfcb4967418bfb8ac142f64a-Paper.pdf}
  {Language models are few-shot learners}.
\newblock In \emph{Advances in Neural Information Processing Systems},
  volume~33, pages 1877--1901. Curran Associates, Inc.

\bibitem[{Chakrabarty et~al.(2020{\natexlab{a}})Chakrabarty, Ghosh, Muresan,
  and Peng}]{chakrabarty-etal-2020-r}
Tuhin Chakrabarty, Debanjan Ghosh, Smaranda Muresan, and Nanyun Peng.
  2020{\natexlab{a}}.
\newblock \href {https://doi.org/10.18653/v1/2020.acl-main.711} {{R}{\^{}}3:
  Reverse, retrieve, and rank for sarcasm generation with commonsense
  knowledge}.
\newblock In \emph{Proceedings of the 58th Annual Meeting of the Association
  for Computational Linguistics}, pages 7976--7986, Online. Association for
  Computational Linguistics.

\bibitem[{Chakrabarty et~al.(2020{\natexlab{b}})Chakrabarty, Muresan, and
  Peng}]{chakrabarty-etal-2020-generating}
Tuhin Chakrabarty, Smaranda Muresan, and Nanyun Peng. 2020{\natexlab{b}}.
\newblock \href {https://doi.org/10.18653/v1/2020.emnlp-main.524} {Generating
  similes effortlessly like a pro: A style transfer approach for simile
  generation}.
\newblock In \emph{Proceedings of the 2020 Conference on Empirical Methods in
  Natural Language Processing (EMNLP)}, pages 6455--6469, Online. Association
  for Computational Linguistics.

\bibitem[{Da et~al.(2021)Da, Bras, Lu, Choi, and Bosselut}]{Da2020AnalyzingCE}
Jeff Da, Ronan~Le Bras, Ximing Lu, Yejin Choi, and Antoine Bosselut. 2021.
\newblock Analyzing commonsense emergence in few-shot knowledge models.
\newblock In \emph{Proceedings of the Conference on Automated Knowledge Base
  Construction (AKBC)}.

\bibitem[{Dambekodi et~al.(2020)Dambekodi, Frazier, Ammanabrolu, and
  Riedl}]{dambekodi-etal-2020}
Sahith Dambekodi, Spencer Frazier, Prithviraj Ammanabrolu, and Mark Riedl.
  2020.
\newblock Playing text-based games with common sense.

\bibitem[{Devlin et~al.(2019)Devlin, Chang, Lee, and
  Toutanova}]{devlin-etal-2019-bert}
Jacob Devlin, Ming-Wei Chang, Kenton Lee, and Kristina Toutanova. 2019.
\newblock \href {https://doi.org/10.18653/v1/N19-1423} {{BERT}: Pre-training of
  deep bidirectional transformers for language understanding}.
\newblock In \emph{Proceedings of the 2019 Conference of the North {A}merican
  Chapter of the Association for Computational Linguistics: Human Language
  Technologies, Volume 1 (Long and Short Papers)}, pages 4171--4186,
  Minneapolis, Minnesota. Association for Computational Linguistics.

\bibitem[{Gao et~al.(2022)Gao, Hwang, Kanno, Wakaki, Mitsufuji, and
  Bosselut}]{silin2022comfact}
Silin Gao, Jena~D. Hwang, Saya Kanno, Hiromi Wakaki, Yuki Mitsufuji, and
  Antoine Bosselut. 2022.
\newblock \href {https://doi.org/10.48550/ARXIV.2210.12678} {Comfact: A
  benchmark for linking contextual commonsense knowledge}.
\newblock In \emph{Findings of EMNLP}.

\bibitem[{Gehman et~al.(2020)Gehman, Gururangan, Sap, Choi, and
  Smith}]{Gehman2020RealToxicityPromptsEN}
Samuel Gehman, Suchin Gururangan, Maarten Sap, Yejin Choi, and Noah~A. Smith.
  2020.
\newblock Realtoxicityprompts: Evaluating neural toxic degeneration in language
  models.
\newblock \emph{ArXiv}, abs/2009.11462.

\bibitem[{Hwang et~al.(2021)Hwang, Bhagavatula, Bras, Da, Sakaguchi, Bosselut,
  and Choi}]{Hwang2021COMETATOMIC2O}
Jena~D. Hwang, Chandra Bhagavatula, Ronan~Le Bras, Jeff Da, Keisuke Sakaguchi,
  Antoine Bosselut, and Yejin Choi. 2021.
\newblock Comet-atomic 2020: On symbolic and neural commonsense knowledge
  graphs.
\newblock In \emph{AAAI}.

\bibitem[{Jiang et~al.(2021)Jiang, Bosselut, Bhagavatula, and
  Choi}]{jiang-etal-2021-im}
Liwei Jiang, Antoine Bosselut, Chandra Bhagavatula, and Yejin Choi. 2021.
\newblock \href {https://doi.org/10.18653/v1/2021.naacl-main.346} {{``}{I}{'}m
  not mad{''}: Commonsense implications of negation and contradiction}.
\newblock In \emph{Proceedings of the 2021 Conference of the North American
  Chapter of the Association for Computational Linguistics: Human Language
  Technologies}, pages 4380--4397, Online. Association for Computational
  Linguistics.

\bibitem[{Kingma and Ba(2015)}]{adam2015}
Diederik~P. Kingma and Jimmy Ba. 2015.
\newblock \href {http://arxiv.org/abs/1412.6980} {Adam: A method for stochastic
  optimization}.
\newblock In \emph{ICLR (Poster)}.

\bibitem[{Lavie and Agarwal(2007)}]{lavie-agarwal-2007-meteor}
Alon Lavie and Abhaya Agarwal. 2007.
\newblock \href {https://aclanthology.org/W07-0734} {{METEOR}: An automatic
  metric for {MT} evaluation with high levels of correlation with human
  judgments}.
\newblock In \emph{Proceedings of the Second Workshop on Statistical Machine
  Translation}, pages 228--231, Prague, Czech Republic. Association for
  Computational Linguistics.

\bibitem[{Lin(2004)}]{lin-2004-rouge}
Chin-Yew Lin. 2004.
\newblock \href {https://aclanthology.org/W04-1013} {{ROUGE}: A package for
  automatic evaluation of summaries}.
\newblock In \emph{Text Summarization Branches Out}, pages 74--81, Barcelona,
  Spain. Association for Computational Linguistics.

\bibitem[{Majumder et~al.(2020)Majumder, Jhamtani, Berg-Kirkpatrick, and
  McAuley}]{majumder-etal-2020-like}
Bodhisattwa~Prasad Majumder, Harsh Jhamtani, Taylor Berg-Kirkpatrick, and
  Julian McAuley. 2020.
\newblock \href {https://doi.org/10.18653/v1/2020.emnlp-main.739} {Like hiking?
  you probably enjoy nature: Persona-grounded dialog with commonsense
  expansions}.
\newblock In \emph{Proceedings of the 2020 Conference on Empirical Methods in
  Natural Language Processing (EMNLP)}, pages 9194--9206, Online. Association
  for Computational Linguistics.

\bibitem[{Papineni et~al.(2002)Papineni, Roukos, Ward, and
  Zhu}]{papineni-etal-2002-bleu}
Kishore Papineni, Salim Roukos, Todd Ward, and Wei-Jing Zhu. 2002.
\newblock \href {https://doi.org/10.3115/1073083.1073135} {{B}leu: a method for
  automatic evaluation of machine translation}.
\newblock In \emph{Proceedings of the 40th Annual Meeting of the Association
  for Computational Linguistics}, pages 311--318, Philadelphia, Pennsylvania,
  USA. Association for Computational Linguistics.

\bibitem[{Pennington et~al.(2014)Pennington, Socher, and
  Manning}]{pennington-etal-2014-glove}
Jeffrey Pennington, Richard Socher, and Christopher Manning. 2014.
\newblock \href {https://doi.org/10.3115/v1/D14-1162} {{G}lo{V}e: Global
  vectors for word representation}.
\newblock In \emph{Proceedings of the 2014 Conference on Empirical Methods in
  Natural Language Processing ({EMNLP})}, pages 1532--1543, Doha, Qatar.
  Association for Computational Linguistics.

\bibitem[{Petroni et~al.(2019)Petroni, Rockt{\"a}schel, Riedel, Lewis, Bakhtin,
  Wu, and Miller}]{petroni-etal-2019-language}
Fabio Petroni, Tim Rockt{\"a}schel, Sebastian Riedel, Patrick Lewis, Anton
  Bakhtin, Yuxiang Wu, and Alexander Miller. 2019.
\newblock \href {https://doi.org/10.18653/v1/D19-1250} {Language models as
  knowledge bases?}
\newblock In \emph{Proceedings of the 2019 Conference on Empirical Methods in
  Natural Language Processing and the 9th International Joint Conference on
  Natural Language Processing (EMNLP-IJCNLP)}, pages 2463--2473, Hong Kong,
  China. Association for Computational Linguistics.

\bibitem[{Radford and Narasimhan(2018)}]{Radford2018ImprovingLU}
Alec Radford and Karthik Narasimhan. 2018.
\newblock Improving language understanding by generative pre-training.

\bibitem[{Radford et~al.(2019)Radford, Wu, Child, Luan, Amodei, and
  Sutskever}]{radford2019language}
Alec Radford, Jeff Wu, Rewon Child, David Luan, Dario Amodei, and Ilya
  Sutskever. 2019.
\newblock Language models are unsupervised multitask learners.

\bibitem[{Roberts et~al.(2020)Roberts, Raffel, and
  Shazeer}]{roberts-etal-2020-much}
Adam Roberts, Colin Raffel, and Noam Shazeer. 2020.
\newblock \href {https://doi.org/10.18653/v1/2020.emnlp-main.437} {How much
  knowledge can you pack into the parameters of a language model?}
\newblock In \emph{Proceedings of the 2020 Conference on Empirical Methods in
  Natural Language Processing (EMNLP)}, pages 5418--5426, Online. Association
  for Computational Linguistics.

\bibitem[{Sanh et~al.(2019)Sanh, Debut, Chaumond, and
  Wolf}]{https://doi.org/10.48550/arxiv.1910.01108}
Victor Sanh, Lysandre Debut, Julien Chaumond, and Thomas Wolf. 2019.
\newblock \href {https://doi.org/10.48550/ARXIV.1910.01108} {Distilbert, a
  distilled version of bert: smaller, faster, cheaper and lighter}.

\bibitem[{Shen et~al.(2018)Shen, Wang, Wang, Min, Su, Zhang, Li, Henao, and
  Carin}]{shen-etal-2018-baseline}
Dinghan Shen, Guoyin Wang, Wenlin Wang, Martin~Renqiang Min, Qinliang Su, Yizhe
  Zhang, Chunyuan Li, Ricardo Henao, and Lawrence Carin. 2018.
\newblock \href {https://doi.org/10.18653/v1/P18-1041} {Baseline needs more
  love: On simple word-embedding-based models and associated pooling
  mechanisms}.
\newblock In \emph{Proceedings of the 56th Annual Meeting of the Association
  for Computational Linguistics (Volume 1: Long Papers)}, pages 440--450,
  Melbourne, Australia. Association for Computational Linguistics.

\bibitem[{Sheng et~al.(2020)Sheng, Chang, Natarajan, and
  Peng}]{sheng-etal-2020-towards}
Emily Sheng, Kai-Wei Chang, Prem Natarajan, and Nanyun Peng. 2020.
\newblock \href {https://doi.org/10.18653/v1/2020.findings-emnlp.291} {Towards
  {C}ontrollable {B}iases in {L}anguage {G}eneration}.
\newblock In \emph{Findings of the Association for Computational Linguistics:
  EMNLP 2020}, pages 3239--3254, Online. Association for Computational
  Linguistics.

\bibitem[{Speer and Havasi(2013)}]{Speer2013ConceptNet5A}
Robyn Speer and Catherine Havasi. 2013.
\newblock Conceptnet 5: A large semantic network for relational knowledge.
\newblock In \emph{The People's Web Meets NLP}.

\bibitem[{Vedantam et~al.(2015)Vedantam, Zitnick, and Parikh}]{7299087}
Ramakrishna Vedantam, C.~Lawrence Zitnick, and Devi Parikh. 2015.
\newblock \href {https://doi.org/10.1109/CVPR.2015.7299087} {Cider:
  Consensus-based image description evaluation}.
\newblock In \emph{2015 IEEE Conference on Computer Vision and Pattern
  Recognition (CVPR)}, pages 4566--4575.

\bibitem[{Wallace et~al.(2019)Wallace, Feng, Kandpal, Gardner, and
  Singh}]{wallace-etal-2019-universal}
Eric Wallace, Shi Feng, Nikhil Kandpal, Matt Gardner, and Sameer Singh. 2019.
\newblock \href {https://doi.org/10.18653/v1/D19-1221} {Universal adversarial
  triggers for attacking and analyzing {NLP}}.
\newblock In \emph{Proceedings of the 2019 Conference on Empirical Methods in
  Natural Language Processing and the 9th International Joint Conference on
  Natural Language Processing (EMNLP-IJCNLP)}, pages 2153--2162, Hong Kong,
  China. Association for Computational Linguistics.

\bibitem[{Weidinger et~al.(2021)Weidinger, Mellor, Rauh, Griffin, Uesato,
  Huang, Cheng, Glaese, Balle, Kasirzadeh, Kenton, Brown, Hawkins, Stepleton,
  Biles, Birhane, Haas, Rimell, Hendricks, Isaac, Legassick, Irving, and
  Gabriel}]{ethical-social-risks}
Laura Weidinger, John Mellor, Maribeth Rauh, Conor Griffin, Jonathan Uesato,
  Po{-}Sen Huang, Myra Cheng, Mia Glaese, Borja Balle, Atoosa Kasirzadeh, Zac
  Kenton, Sasha Brown, Will Hawkins, Tom Stepleton, Courtney Biles, Abeba
  Birhane, Julia Haas, Laura Rimell, Lisa~Anne Hendricks, William~S. Isaac,
  Sean Legassick, Geoffrey Irving, and Iason Gabriel. 2021.
\newblock \href {http://arxiv.org/abs/2112.04359} {Ethical and social risks of
  harm from language models}.
\newblock \emph{CoRR}, abs/2112.04359.

\bibitem[{West et~al.(2022)West, Bhagavatula, Hessel, Hwang, Jiang, Bras, Lu,
  Welleck, and Choi}]{West2022SymbolicKD}
Peter West, Chandrasekhar Bhagavatula, Jack Hessel, Jena~D. Hwang, Liwei Jiang,
  Ronan~Le Bras, Ximing Lu, Sean Welleck, and Yejin Choi. 2022.
\newblock Symbolic knowledge distillation: from general language models to
  commonsense models.
\newblock In \emph{NAACL}.

\bibitem[{Zhang et~al.(2019)Zhang, Kishore, Wu, Weinberger, and
  Artzi}]{https://doi.org/10.48550/arxiv.1904.09675}
Tianyi Zhang, Varsha Kishore, Felix Wu, Kilian~Q. Weinberger, and Yoav Artzi.
  2019.
\newblock \href {https://doi.org/10.48550/ARXIV.1904.09675} {Bertscore:
  Evaluating text generation with bert}.

\end{thebibliography}


\end{document}